\documentclass[conference]{IEEEtran}

\usepackage[utf8]{inputenc}
\usepackage[T1]{fontenc}
\usepackage{csquotes}
\usepackage{amsmath,amssymb}
\usepackage{booktabs}
\usepackage{graphicx}
\usepackage{subcaption}
\usepackage{algorithm}
\usepackage{algorithmic}
\usepackage{xcolor}
\usepackage{xspace}

\usepackage[
  backend=biber,
  bibencoding=utf8,
  style=ieee,
  natbib=true,
  sortcites,
  giveninits=true,
  maxbibnames=3,
  minbibnames=1,
  doi=false,
  url=false,
  isbn=false,
  eprint=false,
  date=year
]{biblatex}
\usepackage[colorlinks=true,linkcolor=blue,citecolor=blue,urlcolor=blue,
            pdfborder={0 0 0}]{hyperref}

\usepackage[printonlyused]{acronym}

\graphicspath{{figs/}}

\newcommand{\ridge}{\textsc{Ridge}\xspace}

\acrodef{PPO}{Proximal Policy Optimization}
\acrodef{GAE}{Generalized Advantage Estimation}
\acrodef{IMPALA}{Importance-Weighted Actor-Learner Architecture}
\acrodef{MORL}{Multi-Objective Reinforcement Learning}
\acrodef{CNN}{Convolutional Neural Network}
\acrodef{LLM}{Large Language Model}

\begin{document}

\title{RIDGE: State-Conditioned Reward Blending\\
       for Behavioral Coverage in Deep RL Game Agents}

\author{\IEEEauthorblockN{AL Muqshith Mohammed Shifan, Kevin Christopher Chua, Ali Neshati,\\Loutfouz Zaman, Cristiano Politowski}
\IEEEauthorblockA{\textit{Ontario Tech University}\\
\{mohamealmusqi.mohamedshifan, kevinchristopher.chua\}@ontariotechu.net\\
\{ali.neshati, loutfouz.zaman, cristiano.politowski\}@ontariotechu.ca}}

\maketitle

\begin{abstract}
A deep RL agent trained on a fixed persona reward plays one way: an
Explorer keeps exploring while starving, a Survivor hoards on an
empty map.  Since each persona fixes its own reward, automated
playtesting trains $N$ agents for $N$ styles.  Changing reward
composition mid-episode runs into the
switching-stability dilemma: \ac{PPO} degrades under non-stationary
rewards.  We present \ridge, a single-agent architecture that
sidesteps the dilemma with two design choices: a state-conditioned
blender that computes softmax-normalized sigmoid weights from
slow-moving state variables, and a multi-head critic that trains
one value head per persona on its own stationary signal.  On Crafter
with four personas, one \ridge agent acquires both branches of the
wood-tier crafting/combat split that no fixed-persona baseline solves jointly, and a
sharpness ablation confirms state-conditioning drives the gain.  We
position these results as demonstrations of the blending mechanism's
dynamics across $n\!=\!3$ seeds per condition; larger multi-seed
sweeps remain future work.
\end{abstract}


\acresetall

\section{Introduction}
\label{sec:intro}

A starving player stops exploring to find food and then resumes once
he is satiated. Human players do this naturally, and automated playtesters
need to cover the same kind of within-episode shift, yet a deep RL
agent trained on a fixed reward plays one way: an Explorer keeps
exploring while starving, a Survivor hoards on an empty map.  Game
developers work around this by defining \emph{procedural
personas}~\cite{holmgard2019}: they pair each agent with a different
reward composition, train one agent per persona, then take the union
of behaviors at deployment.  This scales linearly ($N$ personas cost
$N$ training runs and $N\!\times\!$ environment steps), but it also
misses the within-episode dynamics that motivated the personas in
the first place.  In a Crafter-like open-world survival
setting~\cite{hafner2022crafter}, an Explorer rewards map novelty, a
Survivor health and hunger maintenance, a Craftsman tool-building,
and a Warrior combat engagement, while a real player moves between
all four within a single playthrough.

The natural fix is to change reward composition \emph{during} an
episode, conditioned on agent state.  Prior work avoids this because
in-episode reward shifts break \ac{PPO}.  Li~\cite{li2026switching}
formalizes the problem as the \emph{switching-stability dilemma}:
PPO's value function needs a quasi-stationary signal, so hard reward
switches break convergence.  He documents the effect in
\ac{LLM}-guided satellite scheduling, where near-constant weights
outperform carefully tuned dynamic ones.  Moalla
et~al.~\cite{moalla2024pfo} reach the same conclusion from the
representation side: PPO under non-stationarity loses feature rank
and capacity.  An in-episode persona blender therefore has to shift
rewards in a way that PPO can still handle.

\ridge (Reactive Inter-persona Dynamic Goal Engine) addresses this
gap with two design choices.  \textbf{(i) Smooth blending} computes
softmax-normalized sigmoid weights from slow-moving state variables
(health, hunger, crafting progress, enemy proximity), so the
effective reward varies smoothly between nearby states.
\textbf{(ii) A multi-head critic} keeps one value head per persona.
Each head trains on a stationary per-persona reward signal,
regardless of how the blender weights evolve, which isolates
value-function learning from the shifting mixture.

This paper makes three
contributions.
(1)~A state-conditioned blender coupled with a multi-head critic,
designed to address the switching-stability dilemma in persona-based
playtesting.
(2)~Preliminary Crafter evidence that one \ridge agent acquires both
branches of the wood-tier crafting/combat split that no fixed-persona
baseline solves jointly, with $n\!=\!3$ seeds per condition.
(3)~A sharpness ablation showing that state-conditioning, not the
persona set alone, drives the result, plus a direct comparison to
achievement distillation~\cite{moon2023} in
Section~\ref{sec:discussion}: complementary mechanisms, different
deliverables.

\section{Related Work}
\label{sec:related}

\textbf{Personas for playtesting.}
Holmg{\aa}rd et~al.~\cite{holmgard2019} introduced procedural
personas, and later work~\cite{ariyurek2022} extended them.  Both
encode play styles as utility functions with fixed weights.  Ariyurek
et~al.~\cite{ariyurek2022} argue against static personas and propose
a developing persona whose goals progress over time, together with
an Alternative Path Finder that adjusts the PPO reward for
trajectory diversity.  Their changes act between episodes; \ridge's
act within an episode and condition on state.  Curiosity-driven
coverage agents~\cite{gordillo2021} go after broader behavioral reach
through intrinsic exploration bonuses, not persona blending.  The
closest contemporary works supply the blending signal externally.
MultiGAIL~\cite{ahlberg2023} trains one agent to imitate multiple
persona policies from human demonstrations, conditioned on an
\emph{external} blending coefficient set at inference time.  Cilan
and \"Ozg\"ovde~\cite{cilanControllableDiverseBehaviors2025} condition
a single PPO policy on an external behavior-target vector.  \ridge
computes the blending signal \emph{internally} from game state, so
the mixture reflects behavioral need rather than user prior.

\textbf{Adaptive reward weighting and multi-head critics.}
The closest architectural precedent is HDPG~\cite{huang2023hdpg}, a
multi-head critic for bipedal locomotion where each head estimates a
single reward component and a weight \emph{schedule} (training
progress) combines them.  \ridge uses the same multi-head pattern,
but with semantically interpretable persona heads and weights that
vary with \emph{state within each episode}.  Compared to standard
\ac{MORL}~\cite{mossalam2016,vanmoffaert2014}, which treats
scalarization weights as external configuration, \ridge performs
state-conditioned scalarization with internal weights.  This also
separates \ridge from hard intra-episode switching, whose abrupt
transitions trigger Li's dilemma.

\textbf{PPO non-stationarity.}
Li~\cite{li2026switching} formalizes the switching-stability dilemma:
near-constant reward weights ($342$~Mbps) outperform tuned dynamic
weights ($103$~Mbps) because PPO requires a quasi-stationary signal
for value-function convergence.  Moalla
et~al.~\cite{moalla2024pfo} address the representation-side
counterpart with Proximal Feature Optimization.  \ridge's two design
choices target the value-function side of the same phenomenon:
smooth blending keeps the aggregate reward quasi-stationary, and the
multi-head critic isolates each persona's value estimate from the
shifting mixture.

\textbf{Crafter benchmark methods.}
Several methods target Crafter~\cite{hafner2022crafter} directly,
including achievement distillation~\cite{moon2023}.  Two caveats
matter when comparing absolute scores.  Moon
et~al.~\cite{moon2023} report PPO baselines using an
\ac{IMPALA}~\cite{espeholt2018impala} ResNet encoder ($4$M
parameters) reaching $15.60\%$, far above the $\sim\!4\%$ PPO
Hafner~\cite{hafner2022crafter} reports under a simpler \ac{CNN};
Moon's headline $21.79\%$ uses $9$M parameters and a $1$M-parameter
variant scores $17.07\%$, so the gain reflects both model scale and
the contrastive auxiliary loss.  These methods optimize raw
achievement under Crafter's default sparse reward. \ridge instead
prioritizes broad behavioral coverage within-episode, guided by
designer-authored persona shaping.
Section~\ref{sec:discussion} returns to Moon et~al.\ directly.

\section{Method}
\label{sec:method}

\ridge has three components: designer-authored persona reward
functions, a state-conditioned blender, and a multi-head critic.

\textbf{Persona rewards.}
Let $\mathcal{P}\!=\!\{E,S,C,W\}$ denote Explorer, Survivor,
Craftsman, and Warrior.  Each persona defines a reward
$r_t^i\!:\!(s_t,a_t,s_{t+1})\!\mapsto\!\mathbb{R}$ that scores the
transition from the perspective of its play style.  Explorer rewards
novel-tile discovery with a decaying per-tile bonus.  Craftsman and
Warrior reward newly unlocked crafting and combat achievements
respectively, with magnitudes scaled by tech-tree depth.  Survivor
combines per-step shaping over the vital vector $(h_t, f_t, d_t,
e_t)$ (health, food, drink, energy) with cliff penalties below
per-vital thresholds and event bonuses (e.g., \texttt{wake\_up}).  A
designer can read each $r^i$ and recognize the intended play style.
We release coefficients and full per-persona definitions with the
codebase, along with all training data and trained
checkpoints.\footnote{Code, data, and checkpoints:
\url{https://github.com/Code-SorceryLab/RIDGE}.}

\textbf{State-conditioned blender.}
At each step the blender produces normalized weights
$\tilde{w}_t\in\Delta^{|\mathcal{P}|-1}$ (the simplex of non-negative
weights summing to one) from a small set of
slow-moving state features.  For persona $i$, a scalar activation
$\sigma_t^i\in[0,1]$ summarizes that style's relevance in $s_t$.  The
raw weight is a sigmoid centered at a per-persona bias $b^i$,
\begin{equation}
  w_t^i = \mathrm{sig}\!\bigl(\alpha\,k^i\,(\sigma_t^i-b^i)\bigr),
  \quad
  \tilde{w}_t^i = \frac{w_t^i}{\sum_{j\in\mathcal{P}} w_t^j},
\label{eq:sigmoid}
\end{equation}
with $\mathrm{sig}(x)\!=\!(1+e^{-x})^{-1}$, $\alpha\!>\!0$ a global
\emph{sharpness} shared across personas, and $k^i$ a per-persona
scale.  The activation signals are
$\sigma_t^{E}\!=\!\mathrm{novelty}(s_t)$,
$\sigma_t^{S}\!=\!1-\min(h_t,f_t,d_t,e_t)$,
$\sigma_t^{C}\!=\!\mathrm{craft\_progress}(s_t)$, and
$\sigma_t^{W}\!=\!\mathrm{enemy\_proximity}(s_t)$, each bounded in
$[0,1]$ and increasing when the corresponding style becomes more
relevant.

Two properties matter.  First, $\tilde{w}_t$ varies continuously
with state, so the effective aggregate
$\bar{r}_t\!=\!\sum_i\tilde{w}_t^i r_t^i$ varies smoothly between
nearby states and is quasi-stationary in
Li's~\cite{li2026switching} sense.  Second, \ridge computes weights
internally from state, which distinguishes it from
MultiGAIL-style imitation~\cite{ahlberg2023} and HDPG-style
schedules~\cite{huang2023hdpg}; both supply the blending signal
externally.  The sharpness $\alpha$ controls transition abruptness:
$\alpha\!\to\!0$ collapses to a uniform mixture
($\tilde{w}^i\!=\!1/|\mathcal{P}|$), while $\alpha\!\to\!\infty$
approaches a hard argmax over $\sigma_t^i$ and recovers the
hard-switching regime Li warns against.

\textbf{Multi-head critic and PPO objective.}
A single policy $\pi_\theta(a_t\mid s_t)$ acts in the environment.
On top of a shared encoder we maintain $|\mathcal{P}|$ value heads
$\{V_\phi^i\}$, one per persona, each trained against per-persona
returns $R_t^i=\sum_{k\geq 0}\gamma^k r_{t+k}^i$ via
\ac{GAE}~\cite{schulman2016gae}.  Each $V_\phi^i$
sees a \emph{stationary} reward signal $r^i$ regardless of how
$\tilde{w}_t$ evolves, which isolates value-function learning from
non-stationarity in the aggregate.  This is the failure mode Li
identifies, and the value-side counterpart of Moalla
et~al.'s~\cite{moalla2024pfo} representation collapse.  The PPO
policy update uses a \emph{blended advantage}: we combine
per-persona advantages $A_t^i=R_t^i-V_\phi^i(s_t)$ as
$\bar{A}_t=\sum_i \tilde{w}_t^i A_t^i$ and substitute into the
standard clipped surrogate.  Value heads train with
$\mathcal{L}^{V}(\phi)=\sum_{i}\mathbb{E}_t[(V_\phi^i(s_t)-R_t^i)^2]$.
Algorithm~\ref{alg:ridge} summarizes a training step.

\begin{algorithm}[t]
\caption{\ridge training step}
\label{alg:ridge}
\begin{algorithmic}[1]
\REQUIRE Policy $\pi_\theta$; value heads $\{V_\phi^i\}_{i\in\mathcal{P}}$;
         persona rewards $\{r^i\}$; activations $\{\sigma^i\}$; biases
         $\{b^i\}$; scales $\{k^i\}$; sharpness $\alpha$; PPO clip
         $\epsilon$; optimizer epochs $K$
\STATE Collect rollout $\tau=\{(s_t,a_t,s_{t+1})\}_{t=0}^{T-1}$ with
       $\pi_\theta$
\FOR{$t=0,\ldots,T{-}1$}
  \STATE $w_t^i\!\gets\!\mathrm{sig}(\alpha k^i(\sigma^i(s_t)-b^i))$
         \hfill\COMMENT{Eq.~\ref{eq:sigmoid}}
  \STATE $\tilde{w}_t^i\!\gets\! w_t^i/\!\sum_j w_t^j$;\quad
         $r_t^i\!\gets\! r^i(s_t,a_t,s_{t+1})$
\ENDFOR
\STATE Compute per-persona returns $R_t^i$ and advantages $A_t^i$
       via GAE on each $r^i$ with $V_\phi^i$
\STATE Blend: $\bar{A}_t\!\gets\!\sum_i\tilde{w}_t^i A_t^i$
\FOR{$K$ optimizer epochs}
  \STATE Update $\theta$ on PPO-clipped objective with advantage
         $\bar{A}_t$ and entropy bonus
  \STATE Update $\phi$ on $\mathcal{L}^{V}$ (per-head MSE)
\ENDFOR
\end{algorithmic}
\end{algorithm}

\section{Experimental Setup}
\label{sec:setup}

We evaluate on Crafter~\cite{hafner2022crafter} under the
benchmark's standard $1$M environment-step training budget;
Crafter's $22$ achievements partition into four categories
(collection, crafting, combat, survival) that align with our
persona set.  These achievements are not independent: most sit on a
sequential tech tree, reachable only once prerequisites are met
(e.g.\ Place Table gates both wood-tier tools, which gate stone- and
iron-tier tools), so a low rate on a deep achievement reflects the
whole preceding chain.  We implement PPO~\cite{schulman2017ppo} in PyTorch,
with all conditions sharing encoder, optimizer (Adam), rollout
length, batch size, GAE coefficient, PPO clip, and entropy bonus;
they differ only in reward signal, blender sharpness, and critic
head count.  Conditions: \ridge ($\alpha\!=\!1.0$, four heads);
four single-head fixed-persona baselines
(Explorer/Survivor/Craftsman/Warrior); a sharpness ablation
$\alpha\in\{0.0,0.5,1.5,2.0\}$ on the \ridge architecture.  The
$\alpha\!=\!0.0$ condition collapses to a uniform mixture
($\tilde{w}^i\!=\!1/|\mathcal{P}|$), which is the same four persona
rewards combined linearly with fixed equal weights; this is our
no-state-conditioning ablation, corresponding to the uniform-weight
case of linear-scalarization
MORL~\cite{mossalam2016,vanmoffaert2014}.

\textbf{Seed budget.}  PPO results carry substantial seed
variance~\cite{henderson2018}, so we run $n\!=\!3$ seeds for every
condition: the \ridge default ($\alpha\!=\!1.0$), all sharpness
ablation values ($\alpha\!\in\!\{0.0, 0.5, 1.5, 2.0\}$), and the four
fixed-persona baselines.  The pickaxe/sword pattern holds across
seeds, and the wood-tier coverage effect sizes are large; we treat
these as demonstrations of the blending dynamics rather than
definitive claims, with a larger multi-seed sweep planned for the
extended version.

\section{Results}
\label{sec:results}

\begin{figure}[t]
\centering
\includegraphics[width=0.95\linewidth]{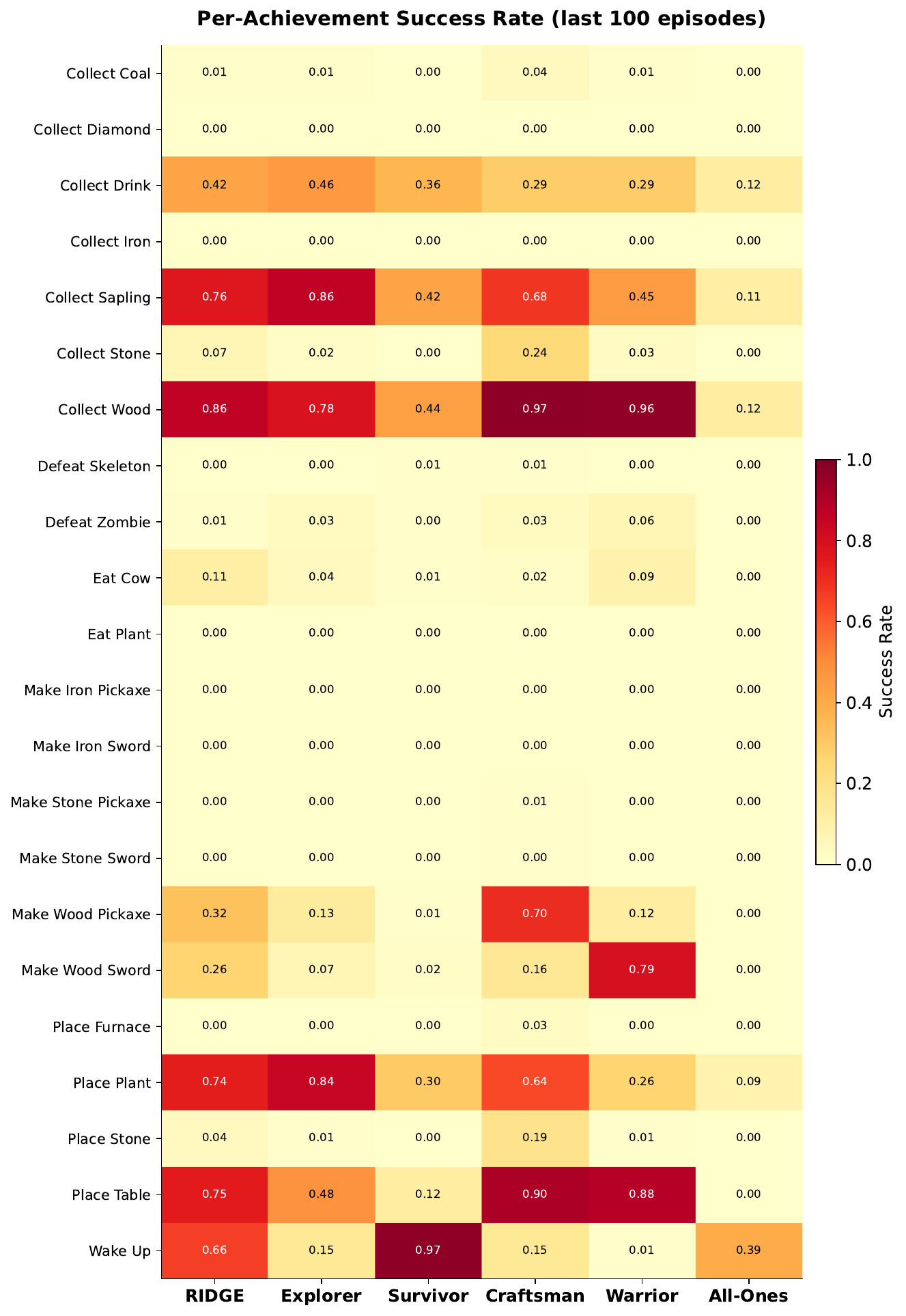}
\caption{Per-achievement success rate over the final $100$ episodes.
\ridge is the only condition with substantial success on both Make
Wood Pickaxe and Make Wood Sword.  \emph{Mean over $n\!=\!3$ seeds
per condition.}}
\label{fig:heatmap}
\end{figure}

\textbf{Cross-branch coverage in a single agent
(Fig.~\ref{fig:heatmap}).}
The clearest result is the wood-tier crafting/combat split.  All
three conditions reach Place Table, the shared prerequisite for both
wood-tier tools: \ridge ($0.75$), Craftsman ($0.90$), Warrior
($0.88$).  After that prerequisite the specialists diverge:
Craftsman makes Wood Pickaxe at $0.70$ but Wood Sword at only
$0.16$, while Warrior is the mirror with $0.79$ on sword and $0.12$
on pickaxe.  \ridge crosses the branch within a single trained agent
and reaches $0.32$ Wood Pickaxe \emph{and} $0.26$ Wood Sword.  Each
specialist's reward gives no incentive for the off-branch skill;
\ridge's composition shifts within episodes, so the agent meets both
incentives in proportion to state.  \ridge sits below each specialist
on its own branch ($0.32$~vs.\ $0.70$ pickaxe, $0.26$~vs.\ $0.79$
sword), the expected coverage/specialization trade-off.  The same
mechanism appears in a smaller Survival case: \ridge reaches Wake~Up
at $0.66$ (vs.\ Explorer $0.15$, Craftsman $0.15$, Warrior $0.01$).
Wake Up fires when the agent sleeps to recover low energy, rewarding
vital management rather than item acquisition, and \ridge reaches it
despite Survivor's mean blender weight sitting around $8\%$, because
the blender spikes Survivor's weight when vitals drop and
policy-gradient signal accumulates over those intervals.

\begin{figure}[t]
\centering
\includegraphics[width=\linewidth]{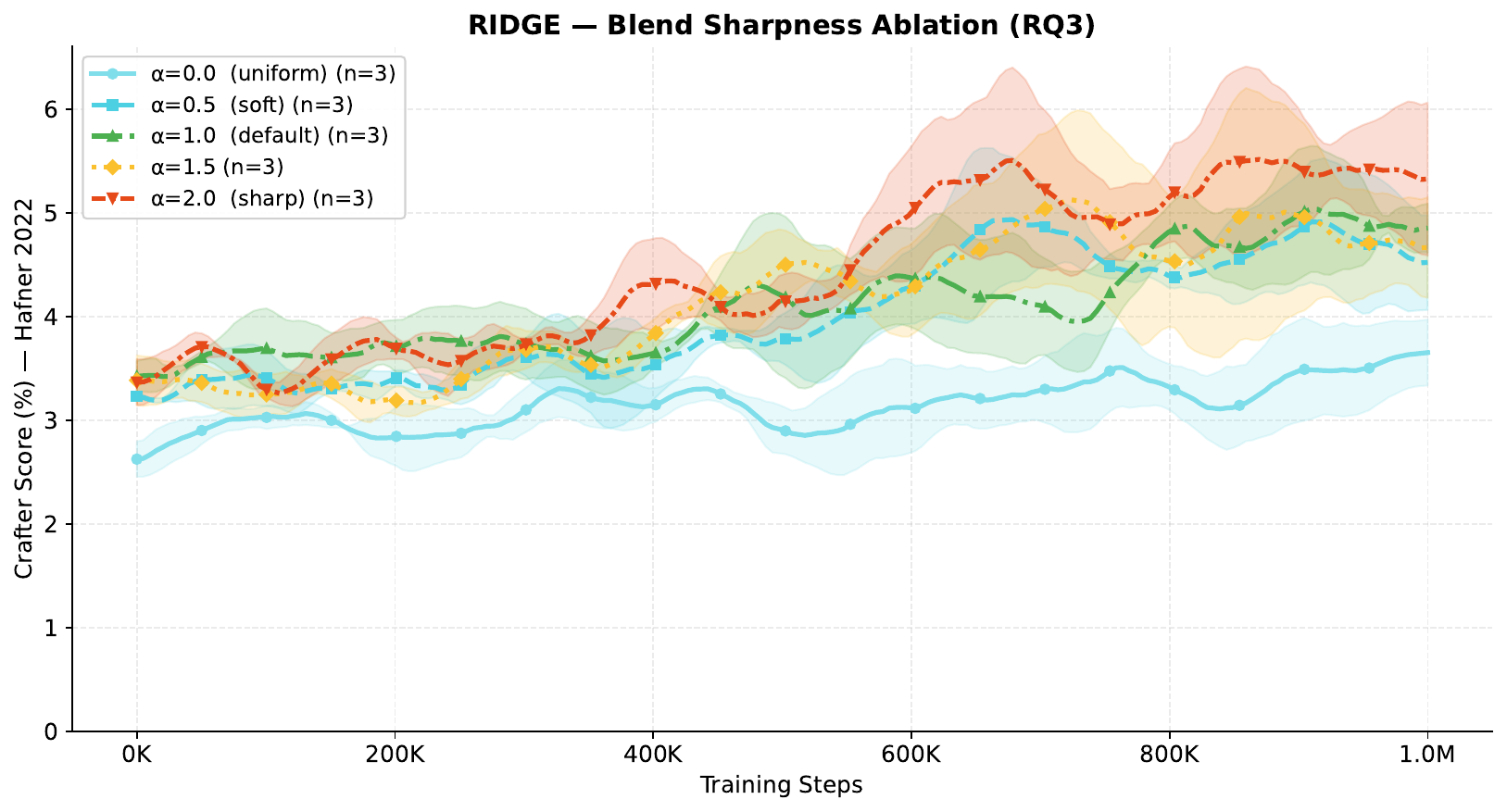}
\caption{Sharpness ablation, Crafter score over training.  Uniform
mixture ($\alpha\!=\!0$) clearly underperforms; state-conditioned
variants cluster well above it.  $n\!=\!3$ seeds for all conditions;
shaded bands are min/max.}
\label{fig:sharpness}
\end{figure}

\textbf{State-conditioning ablation (Fig.~\ref{fig:sharpness}).}
The sharpness sweep tests whether state-conditioning contributes
beyond the persona set itself.  At $\alpha\!=\!0.0$ (uniform mixture,
no state conditioning) the agent receives the same four persona
rewards with equal weights and clearly underperforms all
state-conditioned variants $\alpha\!\in\!\{0.5,1.0,1.5,2.0\}$ by
roughly $30$--$50\%$ relative on aggregate score throughout
training.  Within the conditioned range, the sharper settings
($\alpha\!=\!1.5, 2.0$) edge out the $\alpha\!=\!1.0$ default,
though the differences sit within the seed bands.  Two conclusions
follow.  \emph{First}, state-conditioning, not the persona set
alone, drives the result: the same rewards under uniform weighting
perform worse than under state-conditioned routing.  \emph{Second},
the gap between uniform and state-conditioned conditions is wide
and consistent across seeds; the precise within-conditioned optimum
needs further seeds to pin down.

\begin{figure}[t]
\centering
\includegraphics[width=\linewidth]{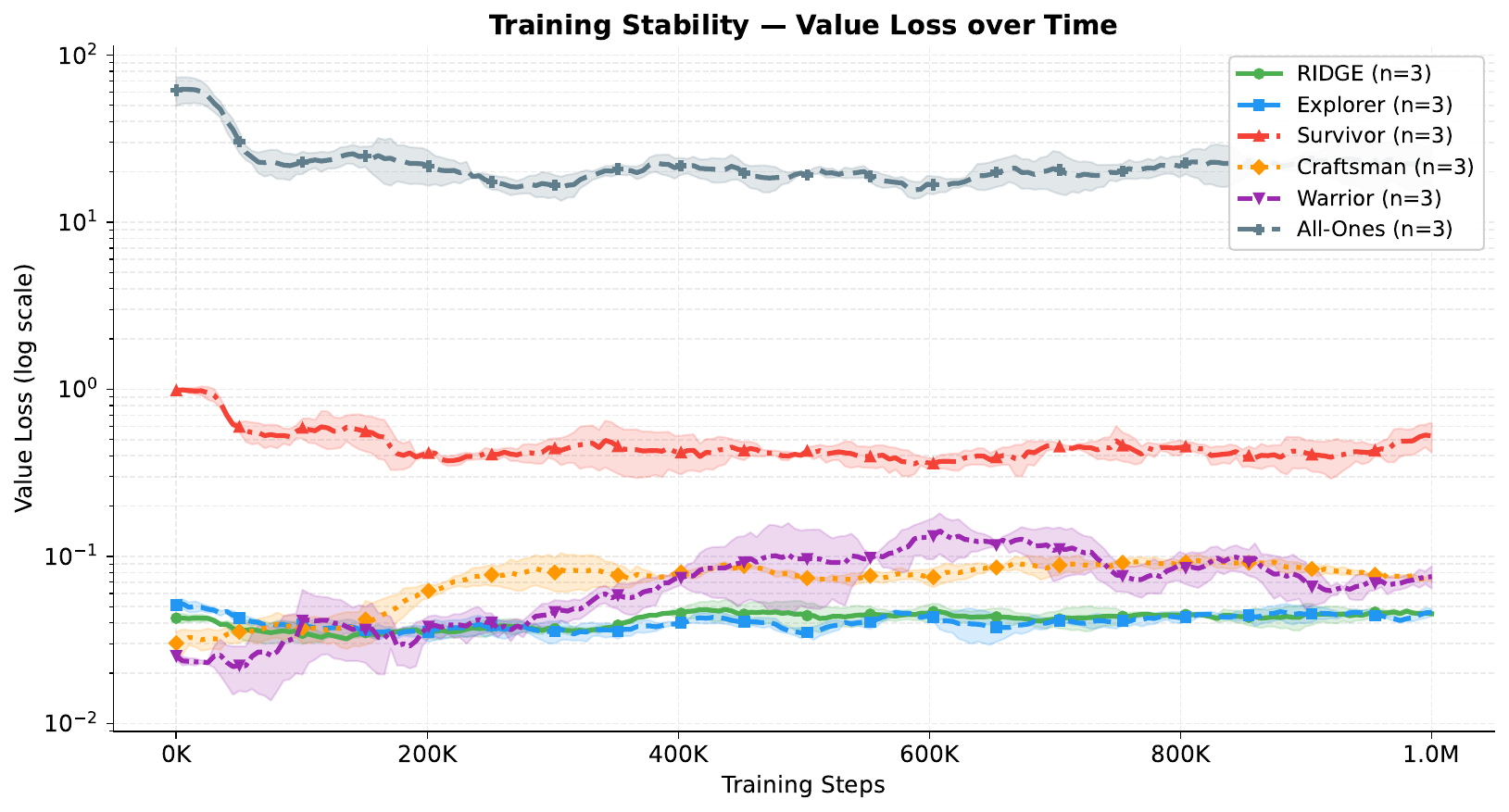}
\caption{Value loss over training (log scale).  \ridge maintains
stable value-function learning under blended rewards, comparable to
the most stable specialist (Explorer).  $n\!=\!3$ seeds per
condition; shaded bands are min/max.}
\label{fig:stability}
\end{figure}

\textbf{Value-function stability (Fig.~\ref{fig:stability}).}
\ridge's value loss stays at $\approx\!0.04$ throughout training,
comparable to Explorer-only and below Craftsman and Warrior.
Survivor-only sits an order of magnitude higher ($0.3$--$0.5$),
because its dense action-independent vital-shaping reward yields
returns the single value head struggles to fit.  With four
per-persona heads each seeing a stationary signal, \ridge's
value-loss trajectory tracks the most stable specialist rather than
degrading toward the non-stationary aggregate, which is direct
evidence for the multi-head critic design choice.

\section{Discussion}
\label{sec:discussion}

\textbf{Relationship to achievement
distillation~\cite{moon2023}.}
The canonical Crafter comparator at $1$M steps is Moon
et~al.~\cite{moon2023}, who reach $21.79\%$ Hafner-canonical score
with $9$M parameters via contrastive auxiliary losses on the
encoder.  \ridge is not on this axis.  We report a $6.30\%$
Hafner-canonical score for reproducibility, not as a competitive
benchmark on this metric.  Moon optimizes raw achievement under
Crafter's default sparse reward, modifying \emph{what the agent
knows}; \ridge optimizes behavioral coverage under designer-authored
persona shaping, modifying \emph{what the agent wants}.  For
automated playtesting, an aggregate achievement score from an agent
locked to one play style provides less value to a designer than a
coverage profile across styles, which is what the wood-tier split
demonstrates.  The two mechanisms are independent in principle and
complementary in practice: contrastive losses on a \ridge-blended
reward, or \ridge blending on a distilled encoder, are both
unexplored.

\section{Conclusion}

\ridge combines a smooth state-conditioned blend over persona rewards
with a critic carrying one value head per persona, letting a single
PPO agent shift between play styles within an episode without breaking
value-function learning. In Crafter, one \ridge agent reaches both the
Wood Pickaxe and Sword, a split no fixed-persona baseline solves, and
a sharpness ablation confirms state-conditioning, not the persona set
alone, drives the gain.

Three extensions would tighten these claims further. While we now report $n\!=\!3$ seeds across all conditions, additional seeds would replace the current min/max bands with confidence intervals. A hard-switch run ($\alpha\rightarrow\infty$) would put Li's dilemma~\cite{li2026switching} under direct test; our $\alpha=2.0$ approaches but does not reach it. Finally, a single-head version of \ridge would separate the multi-head critic's contribution from the smooth blend's contribution. While Crafter is currently the only tested environment, these initial results demonstrate \ridge's potential for training highly adaptable agents.

\renewcommand*{\bibfont}{\scriptsize}
\setlength{\bibitemsep}{0pt}
\printbibliography

\end{document}